\documentclass[10pt,twocolumn,letterpaper]{article}

\usepackage{iccv}
\usepackage{times}
\usepackage{epsfig}
\usepackage{graphicx}
\usepackage{amsmath}
\usepackage{amssymb}
\usepackage{caption}
\usepackage{subcaption}
\usepackage{multirow}
\usepackage{booktabs}
\usepackage{array}

\newcolumntype{C}{>{\centering\arraybackslash}p{4em}}

\usepackage[pagebackref=true,breaklinks=true,letterpaper=true,colorlinks,bookmarks=false]{hyperref}

\iccvfinalcopy 

\def\iccvPaperID{7685} 

\begin{document}

\title{Improving Generalization of Batch Whitening by\\ Convolutional Unit Optimization}

\author{Yooshin Cho ~~~~~~~~~~~~~~~ Hanbyel Cho ~~~~~~~~~~~~~~~ Youngsoo Kim ~~~~~~~~~~~~~~~ Junmo Kim\\
School of Electrical Engineering, KAIST, South Korea \\
{\tt\small \{choys95, tlrl4658, ysoo.kim, junmo.kim\}@kaist.ac.kr}
}

\maketitle

\begin{abstract}

    Batch Whitening is a technique that accelerates and stabilizes training by transforming input features to have a zero mean (Centering) and a unit variance (Scaling), and by removing linear correlation between channels (Decorrelation). In commonly used structures, which are empirically optimized with Batch Normalization, the normalization layer appears between convolution and activation function. Following Batch Whitening studies have employed the same structure without further analysis; even Batch Whitening was analyzed on the premise that the input of a linear layer is whitened. To bridge the gap, we propose a new Convolutional Unit that is in line with the theory, and our method generally improves the performance of Batch Whitening. Moreover, we show the inefficacy of the original Convolutional Unit by investigating rank and correlation of features. As our method is employable off-the-shelf whitening modules, we use Iterative Normalization (IterNorm), the state-of-the-art whitening module, and obtain significantly improved performance on five image classification datasets: CIFAR-10, CIFAR-100, CUB-200-2011, Stanford Dogs, and ImageNet. Notably, we verify that our method improves stability and performance of whitening when using large learning rate, group size, and iteration number. Code is available at \url{https://github.com/YooshinCho/pytorch_ConvUnitOptimization}. 
\end{abstract}

\section{Introduction}
Batch Normalization (BN)~\cite{ioffe2015batch} is considered as a key component of deep neural networks. It significantly stabilizes and accelerates training by normalizing input features to have a zero mean (Centering) and a unit variance (Scaling), which is followed by a linear transform. Numerous follow-up studies have been proposed following the success of BN, and Batch Whitening~\cite{huang2018decorrelated,siarohin2018whitening,huang2019iterative, huang2020investigation, ye2019network} studies were proposed that not only centering and scaling, but also removing linear correlation between input channels (Decorrelation) to improve BN. Although it is well-known that decorrelation increases network capacity, and stabilizes and accelerates training, it is not used in practice due to its ambiguity and complexity. Unlike centering and scaling, decorrelation is not an one-to-one mapping transform, and obtaining the inverse square root of the covariance matrix is computationally complex.

\begin{figure}[t]
    \centering
    \begin{subfigure}[b]{0.11\textwidth}
        \centering
        \includegraphics[width=\textwidth]{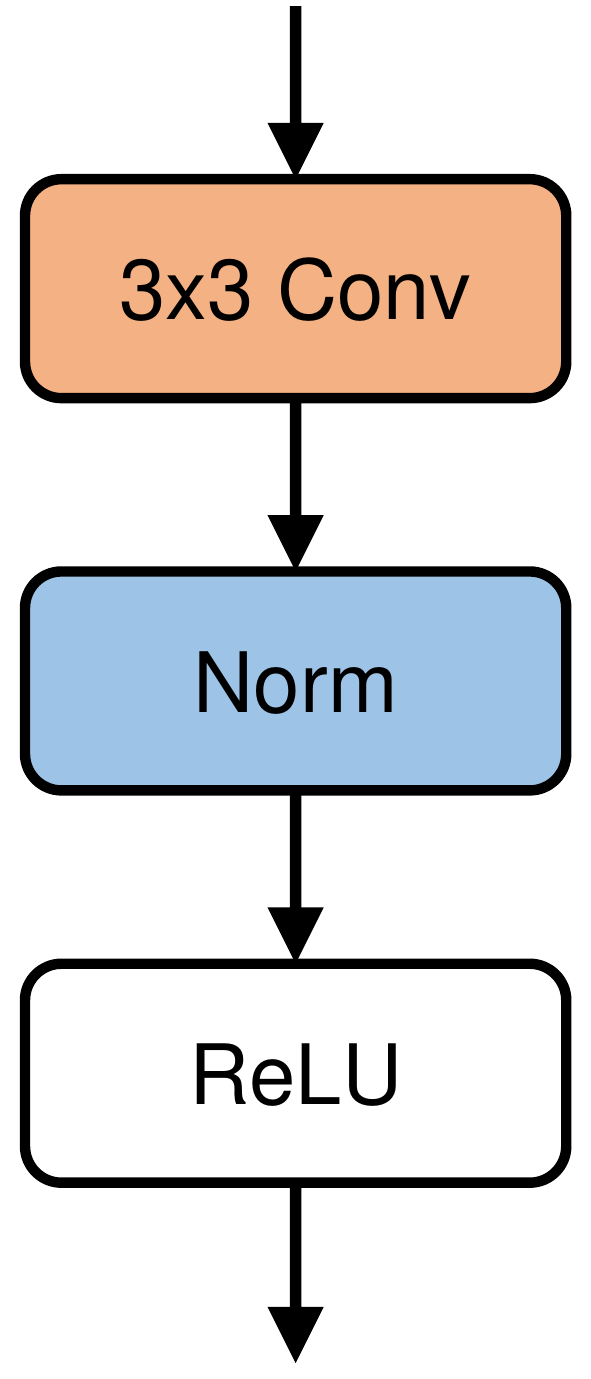}
        \subcaption{Original}
        \label{subfig:curr}
        
    \end{subfigure}
    \hfill
    \begin{subfigure}[b]{0.11\textwidth}
        \centering
        \includegraphics[width=\textwidth]{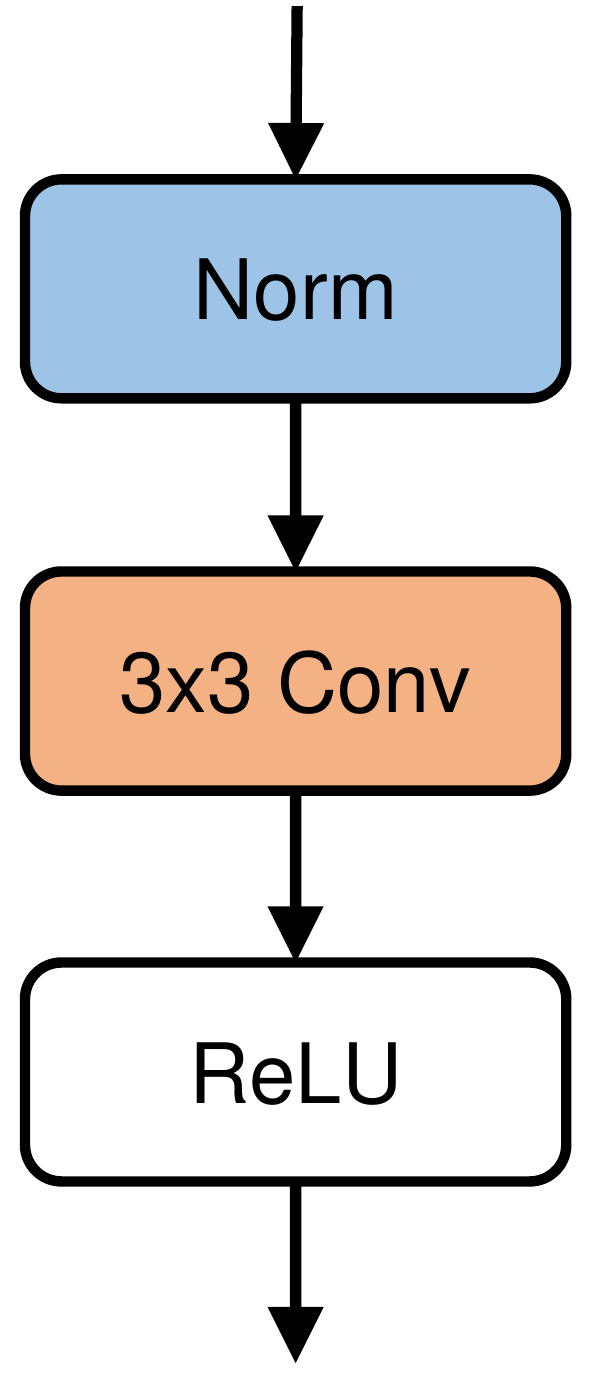}
        \subcaption{Ours}
        \label{subfig:mod1}
        
    \end{subfigure} 
    \hfill
    \begin{subfigure}[b]{0.11\textwidth}
        \centering
        \includegraphics[width=\textwidth]{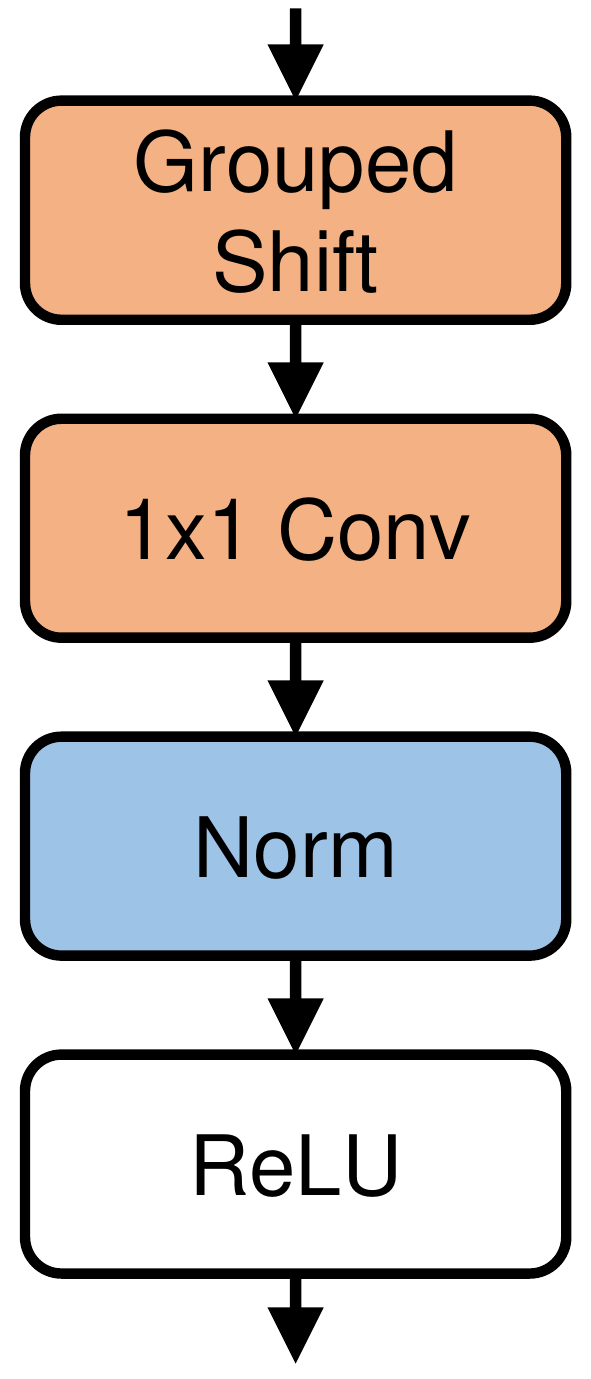}
        \subcaption{Shift}
        \label{subfig:curr2}
        
    \end{subfigure} 
    \hfill
    \begin{subfigure}[b]{0.11\textwidth}
        \centering
        \includegraphics[width=\textwidth]{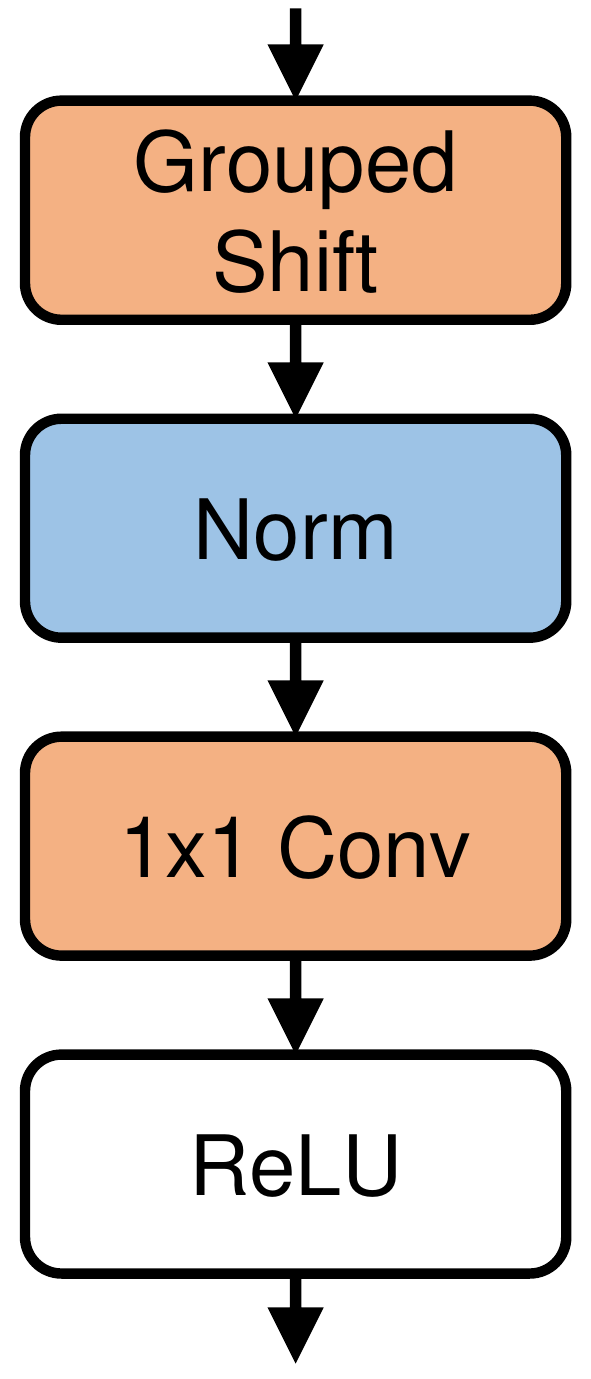}
        \subcaption{Ours+Shift}
        \label{subfig:mod2}
        
    \end{subfigure} 
    \caption{Illustration of the Convolutional Units. (a) is the Original Convolutional Unit. (b) is our modified Convolutional Unit that whitens the input of convolution. (c) is the Original Convolutional Unit that employs Grouped Shift. (d) is our modified Convolutional Unit that employs shift to directly whiten the input of point-wise convolution}
    \vspace{-2mm}
    \label{fig:block}
\end{figure}

Naturally, previous whitening studies have focused on introducing computationally efficient whitening methods, and investigating the reasons for the varying performance of each whitening method. Specifically, whitening methods~\cite{huang2018decorrelated,siarohin2018whitening,huang2019iterative, ye2019network} based on ZCA~\cite{bell1997independent,kessy2018optimal}, Cholesky Decomposition~\cite{dereniowski2003cholesky}, and Newton's iterations~\cite{bini2005algorithms} were proposed, and Iterative Normalization (IterNorm)~\cite{huang2019iterative}, based on Newton's iterations, archeived the state-of-the-art performance owing to its small stochasticity. The superiority of IterNorm was investigated in~\cite{huang2020investigation} by comparing it with other whitening modules, but the performance gain is yet to be fully explored. To investigate the efficacy of IterNorm, we train ResNet~\cite{he2016deep} and Wide-Residual Network (WRN)~\cite{zagoruyko2016wide} on CIFAR-10, CIFAR-100~\cite{krizhevsky2009learning}, and ImageNet~\cite{ILSVRC15}. We apply IterNorm by replacing all BN of ResNet and WRN with IterNorm. From the results shown in Table~\ref{tab:BNvsIN}, we can observe that the performance of IterNorm is not satisfactory on CIFAR-100, despite the correlation of features being successfully removed. 

To identify the reasons for the poor results, we revisit the theory of Batch Whitening~\cite{lecun2012efficient}. We identify the gap between the theory and practice in terms of block design, and assume that the inefficacy of IterNorm can be attributed to the way in which whitening modules are used. Mechanisms of Batch Whitening were analyzed on the premise that the input of the linear layer is whitened; however, in commonly used block design, normalization layer is followed by a linear transform and the activation function before convolution as illustrated in Figure~\ref{subfig:curr}. In this paper, we call these specific order of the layers as ``Convolutional Unit''. This Convolutional Unit was empirically optimized with BN without any analysis, but following Batch Whitening studies~\cite{huang2018decorrelated, siarohin2018whitening, huang2019iterative} have employed the same Convolutional Unit. Thus, irrespective of the efficacy of the whitening modules, the input of convolution is not centered, scaled, and decorrelated. Moreover, there are differences between spatial convolution and the linear layer considered in theory. Spatial convolution can be divided into shift operation~\cite{wu2018shift, jeon2018constructing, chen2019all} and point-wise convolution; thus, there is spatial misalignment between the input of spatial convolution and point-wise convolution. We call the gap as \textit{input misalignment} in this paper. It means the whitening process is affected by the spatial shift operation, even if we directly perform whitening at the input of spatial convolution. 

In this paper, we highlight three structural problems that contributes to the gap between the theory and practice: linear transform, position of whitening module, and \textit{input misalignment}. To bridge the gap one step at a time, we modify the Convolutional Unit as illustrated in Figure~\ref{subfig:mod1}. Then, we empirically analyze the original and our Convolutional Unit. We employ IterNorm to empirically confirm the efficacy of whitening module is increased when used with our Convolutional Unit. Series of ablation studies show that the original Convolutional Unit is well-optimized with BN, but not optimized with whitening modules. As we expected, IterNorm outperforms when using with our Convolutional Unit, and linear transform makes training unstable and over-fit. To support superiority of our method, we also investigate the rank and correlation of features. We empirically confirm that correlation is increased by about five times due to linear transform and activation function in practice. Also, we verify that input feature of normalization layer is not full rank when using the original Convolutional Unit. It makes decorrelated output extremely unstable due to noisy channels, and performance degenerate. By contrast, we observe that the input feature of normalization layer is full rank when using our Convolution Unit.

For further improvements, we close the gap, \textit{input misalignment}, which is caused by spatial shift of convolution. To directly perform whitening at the input of point-wise convolution, we divide spatial convolution into Grouped Shift~\cite{wu2018shift} and point-wise convolution, and place whitening modules between them as illustrated at Figure~\ref{subfig:mod2}. With modifications, we get much better performance than BN and IterNorm with original Convolutional Unit on CIFAR-10, CIFAR-100, CUB-200-2011~\cite{WahCUB_200_2011}, Stanford Dogs~\cite{KhoslaYaoJayadevaprakashFeiFei_FGVC2011}. To the best of our knowledge, this is the first paper that shows applicability of whitening modules in transfer learning, which we demonstrate on CUB-200-2011 and Stanford Dogs. Furthermore, we empirically confirm that our method enhances training stability of whitening modules. Our method shows significantly improved results at larger learning rates when the performance of BN, IterNorm using the original Convolution Unit decreases. Also, we compare stability of IterNorm with the original Convolutional Unit and our Convolutional Unit as increasing the iteration number. IterNorm using our Convolutional Unit shows much more stable behavior and significantly better performance with a iteration number larger than 7. Finally, we additionally adopt DBN~\cite{huang2018decorrelated}, and demonstrate that our Convolutional Unit generally improves efficacy of whitening modules. 

\section{Related Works}
\label{sec:related}
\subsection{Batch Standardization}
\label{subsec:batchstand}
Batch Standardization is a technique that only transforms features to have a zero mean (Centering) and a unit variance (Scaling) for computational efficiency. Since the success of Batch Normalization (BN)~\cite{ioffe2015batch}, a number of studies~\cite{DBLP:journals/corr/UlyanovVL16,ba2016layer,Wu_2018_ECCV} have been proposed to improve the speed of learning by standardizing features. These studies focused on improving performance of micro-batch training and fixing a discrepancy between training and inference. They differed only in the target of normalization (e.g. batch, layer, instance, and group of channels), and they all performed the basic procedures of normalization: centering, scaling, and applying linear transformation. Therefore, they can be generally expressed using the following formula:
\begin{equation}
    f(x_{i}) = \gamma_{i} \frac{x_{i}-\mu_{i}}{\sqrt{\sigma^2_{i} + \epsilon}} + \beta_{i}  ,
\label{eq:norm}
\end{equation}
where $\mu$ and $\sigma$ are mean and standard deviation, respectively. The linear transformation is composed by scaling $\gamma$, and shifting $\beta$ as described at Eq~\ref{eq:norm}. It was intuitively, empirically introduced to prevent lose the original representation that can be lost during normalization. Also, the position of normalization was decided to be right after convolution based on the intuition that the output of convolution has ``More-Gaussian'' distribution without further analysis~\cite{ioffe2015batch}. Naturally, the true reasons why BN helps optimization are still an active area of research, and Santurkar~\etal\cite{santurkar2018does} argues that BN helps training by smoothening loss landscape, not by the internal covariance shift.

\subsection{Batch Whitening}
\label{subsec:batchdecorr}
It is generally known that performing centering, scaling, and removing the linear correlations between channel features on the batch helps efficient gradient descent. It conditions the Hessian of the weight and makes first-order gradient descent closer to the second-order gradient descent~\cite{lecun2012efficient}. However, decorrelation is not adopted by BN due to its ambiguity and complexity. Thus, many studies~\cite{huang2018decorrelated,siarohin2018whitening,huang2019iterative,ye2019network} have proposed computationally efficient whitening methods and investigated the reason why whitening modules show different performances~\cite{huang2020investigation}. Specifically, Batch Whitening can be generally expressed as follows,
\begin{equation}
    \boldsymbol{\hat{X}} = \boldsymbol{\Sigma^{-\frac{1}{2}}}\cdot (\boldsymbol{X} - \mu \cdot \boldsymbol{1} ), 
    \label{eq:whitening}
\end{equation}
where $\boldsymbol{X}$ is the input, and $\boldsymbol{\Sigma}$ and $\mu$ are covariance matrix and mean of the input batch, respectively. Decorrelated Batch Normalization (DBN)~\cite{huang2018decorrelated} proposed ZCA whitening to get $\Sigma^{-\frac{1}{2}}$ and addressed the \textit{stochastic axis swapping} issues. \textit{Stochastic axis swapping} is caused by the ambiguity of rotation matrix of PCA whitening~\cite{desjardins2015natural}, but ZCA whitening fixes it by minimizing distortion caused by whitening. Siarohin~\etal\cite{siarohin2018whitening} adopted whitening based on Cholesky Decomposition~\cite{dereniowski2003cholesky} and introduced conditional Coloring transform to improve performance of GAN networks. Iterative Normalization (IterNorm)~\cite{huang2019iterative} proposed whitening based on Newton's iterations~\cite{bini2005algorithms} and shows the state-of-the-art performance. Unlike previous methods, IterNorm approximates $\Sigma^{-\frac{1}{2}}$, and shows the smallest stochasticity.

While there have been many advances on Batch Whitening, there has been no analysis from the structural point of view. Thus, Batch Whitening studies followed the same Convolutional Unit that empirically optimized with BN. In this paper, we analyze the effects of block design, and maximize the efficacy of Batch Whitening by optimizing the Convolutional Unit.


\section{Preliminary}
\label{sec:preli}
In this section, we briefly describe Iterative Normalization and Shift Operation.

\subsection{Iterative Normalization}
\label{subsec:IN}

Iterative Normalization (IterNorm)~\cite{huang2019iterative} is a state-of-the-art Batch Whitening module that employs Netwon's iterations~\cite{bini2005algorithms} to obtain the inverse square root of covariance with controlling stochasticity. From the eq~\ref{eq:whitening}, we can say $\boldsymbol{\hat{X}}$ is a random variable which shows stochasticity caused by the batch sampling~\cite{teye2018bayesian}, and Huang~\etal\cite{huang2020investigation} suggested that the performance of the whitening module is related to its inherent stochasticity. Unlike previous whitening methods, the stochasticity of IterNorm can be controlled by two factors, iteration number and group size. IterNorm approximates the inverse square root of covariance by using an iterative process, and progressively stretches or squeezes the data along the axis to make eigenvalues 1. Also, IterNorm performs group-wise whitening to reduce stochasticity. Therefore, IterNorm performs whitening worse when iteration number and group size are small, but stochasticity is also decreased by ignoring the data along the axis with relatively small eigenvalues. It makes training stable and performance improve, but optimizing the trade-off between stochasticity and degree of whitening is difficult. In this study, we show that stability of IterNorm can be enhanced by Convolutional Unit optimization without loss of capability of whitening.

\subsection{Shift Operation}
The shift operation was originally introduced to reduce the number of parameters and FLOPs by replacing spatial convolution~\cite{wu2018shift,jeon2018constructing,chen2019all}. It was inspired by the fact that spatial convolution can be divided into shift operation and point-wise convolution. Basic spatial convolution can be expressed in the following formula:
    \begin{equation}
    \begin{aligned}
    \label{eq:conv eq matrix form}
    \boldsymbol{Y} = \boldsymbol{\tilde{W}} \times \boldsymbol{\tilde{X}} 
    &= \sum_k {\boldsymbol{W}_{:,:,k} \cdot \boldsymbol{X}^k_{:,:}} \\ 
    &= \sum_k {\boldsymbol{W}_{:,:,k} \cdot S_k(\boldsymbol{X})}, 
    \end{aligned}
    \end{equation}
where $\boldsymbol{\tilde{W}}\in \mathbb{R}^{C_{out}\times kC_{in}}$ and $\boldsymbol{\tilde{X}}\in \mathbb{R}^{kC_{in} \times BHW}$ are weight matrix and concatenation of spatially shifted input matrix, respectively. $k$ is kernel index and $B$ is batch size. $X^k$ is spatially shifted input corresponding to $W_{:,:,k}$, weight of specific kernel index $k$. $S_k(\cdot)$ is a shift operation of displacement of kernel index $k$. 
To directly perform whitening to the input of point-wise convolution, we employ Grouped Shift~\cite{wu2018shift}. Grouped Shift is the operation that spatially shifts the features with fixed displacements. For consistency, we employ ShiftResNet and ShiftNet-A that proposed in~\cite{wu2018shift} as baselines in Section~\ref{subsec:4.shift}. To effectively utilize Grouped Shift operation, channel sizes of intermediate features should be large. Similar to the bottleneck block used in ResNet, ShiftResNet and ShiftNet-A use the ``Expansion Rate'' $\varepsilon$ to control the channel sizes. We conduct experiments with an expansion rate of 6 in the following sections.


\begin{table}[t]
\begin{center}
\small
\begin{tabular}{l|c|c}
\toprule
Dataset / Arch. & BN & IterNorm \\
\toprule
CIFAR-10 / ResNet20 & \textbf{8.18$\pm$0.15} & \textbf{8.17$\pm$0.19} \\
\midrule
CIFAR-10 / WRN-28-10 & 3.76$\pm$0.13 & \textbf{3.68$\pm$0.16} \\
\midrule
CIFAR-100 / ResNet56 & \textbf{27.06$\pm$0.39} & 27.53$\pm$0.35 \\
\midrule
CIFAR-100 / WRN-28-10 & \textbf{18.71$\pm$0.13} & 19.01$\pm$0.20 \\
\midrule
\multirow{2}{*}{ImageNet / ResNet18} & \multirow{2}{*}{29.33} & 29.48 \\
&&\textbf{28.86}\\
\bottomrule
\end{tabular}
\end{center}
\vspace{-3mm}
\caption{Comparison of top-1 test errors (\%) on ResNet and Wide-Residual Network (WRN) on CIFAR-10, CIFAR-100, and ImageNet. Except ImageNet, results are shown in the format of ``mean±std''. For ImageNet, the second row of IterNorm uses the ``Full$+$DF''~\cite{huang2019iterative}. ``Full$+$DF'' means additonal IterNorm is plugged in after the last average pooling. }
\label{tab:BNvsIN}
\end{table}

\begin{table*}[t]
\centering
\small
\begin{subtable}[ht]{0.8\textwidth}
\centering
  \begin{tabular}{l|c|c|c|c}
    \toprule
    Methods & $\gamma,\beta$ & $\gamma$ & $\beta$ & None \\
    \toprule
    BN / Orignial& \textbf{8.18$\pm$0.15} & 8.41$\pm$ 0.22& 8.40$\pm$0.21 & 8.88$\pm$0.20 \\
    \midrule
    \multirow{2}{*}{BN / Ours} & \textbf{8.43$\pm$0.19}& 8.55$\pm$0.19 & 8.68$\pm$0.14& 8.71$\pm$0.29 \\
    &(+0.25) &(+0.14)&(+0.28)&\textbf{(-0.17)}\\
    \toprule
    IterNorm / Orignial & \textbf{8.17$\pm$0.19} & 8.64$\pm$0.22 & 8.26$\pm$0.19 & 9.01$\pm$0.19 \\
    \midrule
    \multirow{2}{*}{IterNorm / Ours} & \textbf{8.17$\pm$0.17} & 8.38$\pm$0.20 & \textbf{8.16$\pm$0.17} & 8.29$\pm$0.20 \\
    &(-0.0)&(-0.26)&(-0.1)&\textbf{(-0.72)}\\
    \bottomrule
    \end{tabular}
    \subcaption{CIFAR-10, ResNet20}
\end{subtable}

\begin{subtable}[ht]{0.8\textwidth}
\centering
  \begin{tabular}{l|c|c|c|c}
    \toprule
    Methods & $\gamma,\beta$ & $\gamma$ & $\beta$ & None \\
    \toprule
    BN / Orignial & \textbf{27.06$\pm$0.40} & 27.49$\pm$0.39 & 27.48$\pm$0.33 & 28.31$\pm$0.24 \\
    \midrule
    \multirow{2}{*}{BN / Ours} & 27.82$\pm$0.30 & 27.82$\pm$0.25 & 27.85$\pm$0.31 & \textbf{27.48$\pm$0.23} \\
    & (+0.76) &(+0.33)&(+0.27)&\textbf{(-0.83)} \\
    \toprule
    IterNorm / Orignial & \textbf{27.53$\pm$0.35} & 28.33$\pm$0.24 & \textbf{27.48$\pm$0.32} & 28.35$\pm$0.37 \\
    \midrule
    \multirow{2}{*}{IterNorm / Ours} & \textbf{27.1$\pm$0.30} & 27.64$\pm$0.30 & \textbf{27.12$\pm$0.25} & 27.3$\pm$0.34 \\
    &(-0.43)&(-0.69)&(-0.36)&\textbf{(-1.05)} \\
    \bottomrule
    \end{tabular}
    \subcaption{CIFAR-100, ResNet56}
\end{subtable}
\vspace{-1mm}
\caption{Comparison of test errors (\%) on ResNet20 and ResNet56 with CIFAR-10/100. All results are computed over 10 random seeds, and shown in the format of ``mean$\pm$std''. The values in parentheses indicate the test error difference between original and proposed unit.}
\label{tab:abl of base,ours}
\end{table*}

\section{Convolutional Unit Optimization}

In this section, we discuss the way in which the Convolutional Unit can be optimized to match theory and to show generally improved performance in practice. We analyze the inefficacy of the linear transform and the position of whitening modules in Section~\ref{sec:lineartransform}, and address \textit{input misalignment} and shift operation in Section~\ref{subsec:4.shift}. To empirically verify efficacy of our method, we employ IterNorm~\cite{huang2019iterative}, the state-of-the-art whitening module. IterNorm is applied by replacing all BN~\cite{ioffe2015batch} with IterNorm with a iteration number of 5 and full group size as suggested in \cite{huang2019iterative}. For our modified Convolutional Unit, we do not use the linear transform in the experiments, unless otherwise stated.

\subsection{Linear Transform and Position of Whitening}
\label{sec:lineartransform}

As shown in Table~\ref{tab:BNvsIN}, we empirically verify the inefficacy of IterNorm, and assume that the sub-optimality of the Convolutional Unit degenerates the performance of the whitening modules. The linear transform following normalization was intuitively introduced to prevent possible loss of representation capability, and the position of normalization layer was decided based on the intuition that the output of convolution is more likely to have a symmetric, non-sparse distribution than the output of activation function without any analysis~\cite{ioffe2015batch}. Based on the premise of theory that the input of the linear layer is whitened, we assume that whitening modules should be placed right before convolution without linear transform. Therefore, we modify the Convolutional Unit as illustrated in Figure~\ref{subfig:mod1} and apply to ResNet. We apply our Convolutional Unit by arranging the position of all normalization layers to be right before convolution.

\begin{figure}[t]
    \centering
    \begin{subfigure}[b]{0.23\textwidth}
        \centering
        \includegraphics[width=\textwidth]{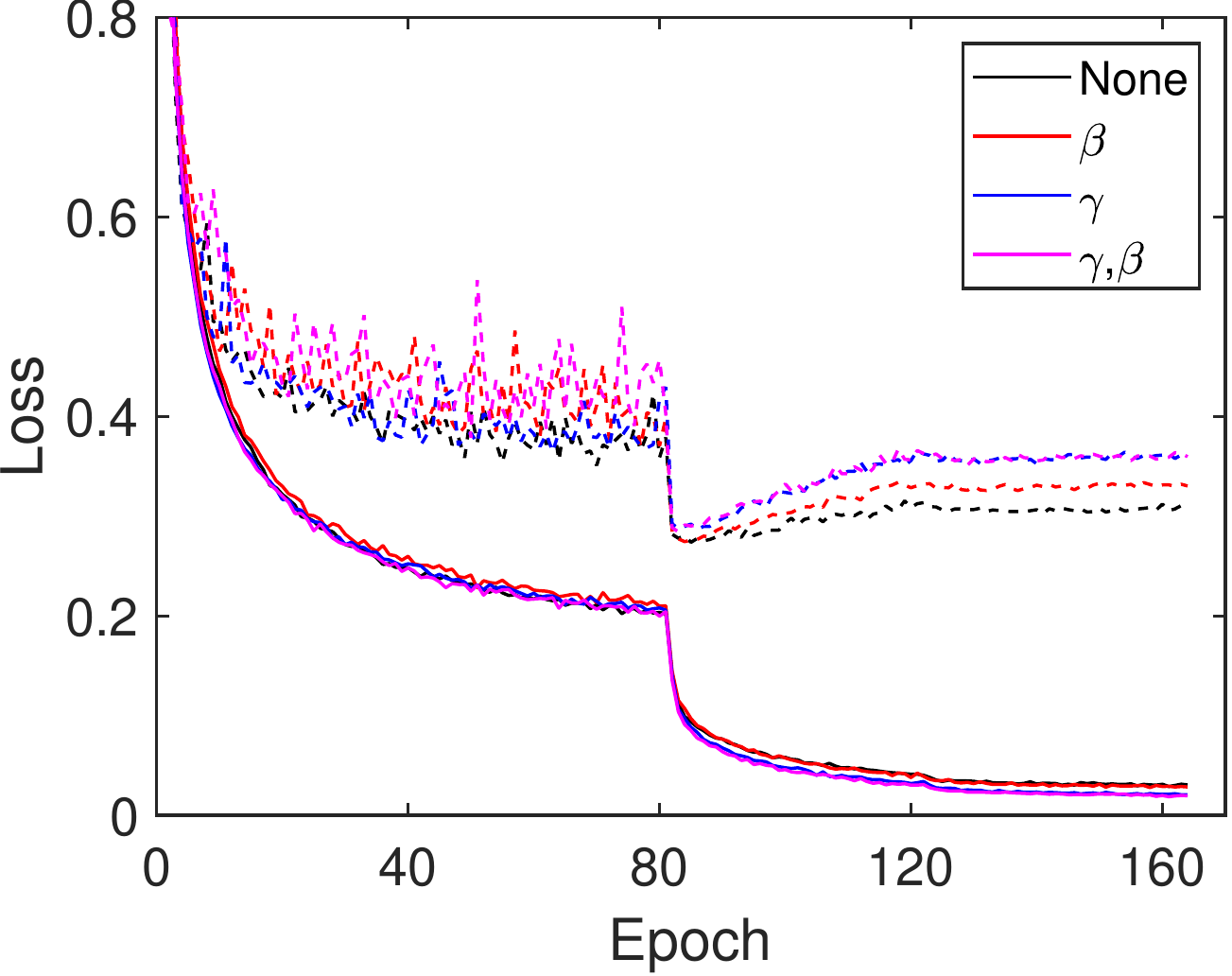}
        \subcaption{CIFAR-10, ResNet20}
    \end{subfigure}
    \hfill
    \begin{subfigure}[b]{0.23\textwidth}
        \centering
        \includegraphics[width=\textwidth]{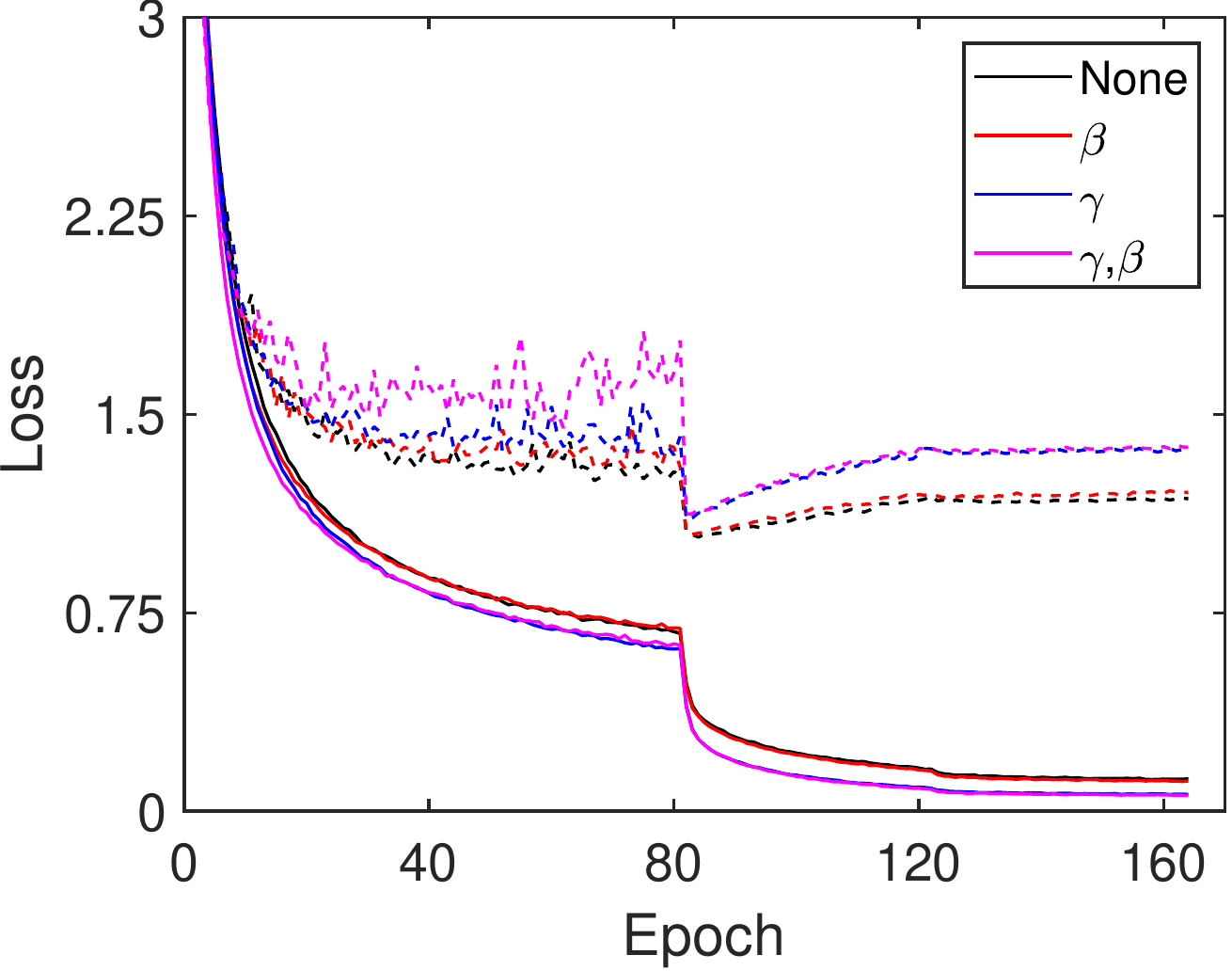}
         \subcaption{CIFAR-100, ResNet56}
    \end{subfigure} 
    \caption{Illustration of train (solid lines), test (dashed lines) loss with respect to epochs. We train ResNet20 and ResNet56 using our Convolutional Unit with IterNorm on CIFAR-10 / 100. It shows the results of ablation studies by varying linear transform. Both (a) and (b) show that $\gamma$ makes learning unstable and over-fit.}
    \vspace{-2mm}
    \label{fig:abl_lineartransform}
\end{figure}

To validate our assumption, we execute ablation studies by varying the Convolutional Units and the linear transform. We denote scaling and shifting operation of the linear transform as $\gamma$ and $\beta$, respectively. We conduct experiments with ResNet~\cite{he2016deep} on two benchmark image classification datasets, CIFAR-10/100~\cite{krizhevsky2009learning}. Table~\ref{tab:abl of base,ours} shows the results of the ablation studies. From the results, we validate that the original Convolutional Unit with linear transform is well-optimized with BN, but not optimized with IterNorm. Performance of IterNorm is generally increased as the Convolutional Unit is modified. Especially, IterNorm without any linear transform shows the largest performance gain when the Convolutional Unit is changed. Although, IterNorm with our Convolutional Unit still shows worse results than BN with original Convolutional Unit, it can be attributed to \textit{input misalignment} which will be addressed in Section~\ref{subsec:4.shift}. Also, in Table~\ref{tab:abl of base,ours}, we can see that $\gamma$ degrades the performance of IterNorm when using our Convolutional Unit. It has shown that removing correlation of input leads to better conditioning of the Hessian in updating the weights, and makes training closer to Newton's method~\cite{lecun2012efficient}. Therefore, scaling factor $\gamma$ is unnecessary when properly utilizing whitening modules. As shown in Figure~\ref{fig:abl_lineartransform}, we can confirm $\gamma$ makes learning unstable, and less compatible with the goal of conditioning Hessian via Batch Whitening.

To demonstrate the benefits of our Convolutional Unit, we investigate correlation of input feature of convolution and rank of input feature of normalization layer. We calculate the mean of the correlation $\rho$, and the mean of the rank divided by channel-size $r$, as follows:
\begin{equation}
    \rho =\frac{1}{L-1} \sum_{l=1}^{L} \frac{2}{C^l(C^l-1)}\sum_{i=0}^{C^l} \sum_{j=i+1}^{C^l}(\boldsymbol{\tilde{X}}^l\boldsymbol{\tilde{X}}^{l\top})_{i,j},
\end{equation}
\begin{equation}
    r =\frac{1}{L-1} \sum_{l=1}^L(\frac{rank(\boldsymbol{X}^l)}{C^l}),
\end{equation}
where $\boldsymbol{\tilde{X^l}},\boldsymbol{X^l} \in \mathbb{R}^{C^l \times BH^lW^l}$ are input matrix of $l$th convolution layer that normalized by $l_2$ norm of each channel and input matrix of $l$th normalization layer, respectively. $L$ and $C^l$ are number of convolution layers and channel size of $\boldsymbol{X^l}$, respectively. We empirically show the severe effects of linear transform and activation function on decorrelation in Figure~\ref{subfig:rank,corr}. Correlation increases by almost five times due to the linear transform and activation function. It indicates efficacy of Batch Whitening is severely affected by the linear transform and the activation function. 

\begin{figure}
    \centering
    \begin{subfigure}[b]{0.23\textwidth}
        \centering
        \includegraphics[width=\textwidth]{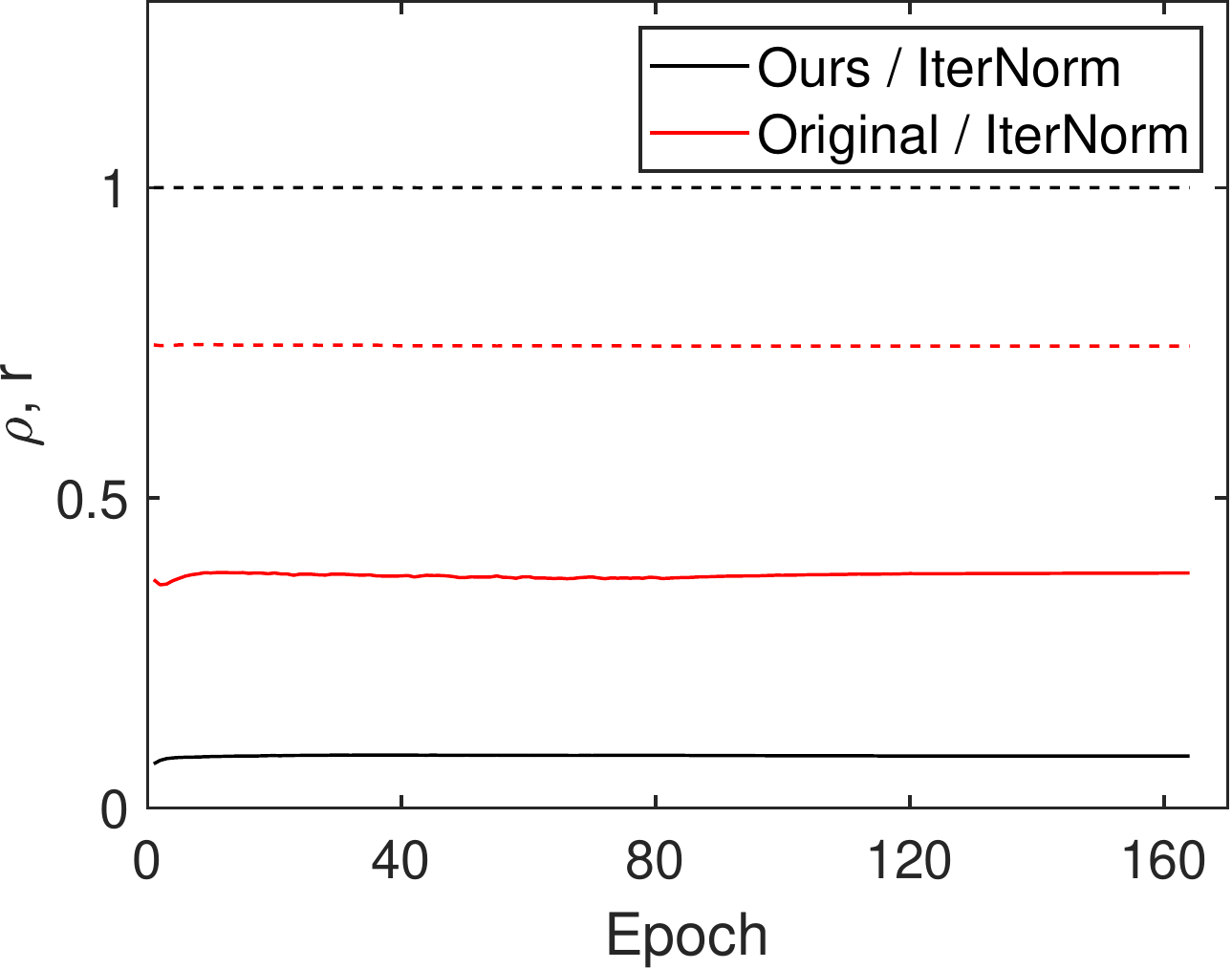}
         \subcaption{Correlation, Rank}
         \label{subfig:rank,corr}
    \end{subfigure} 
    \hfill
    \begin{subfigure}[b]{0.23\textwidth}
        \centering
        \includegraphics[width=\textwidth]{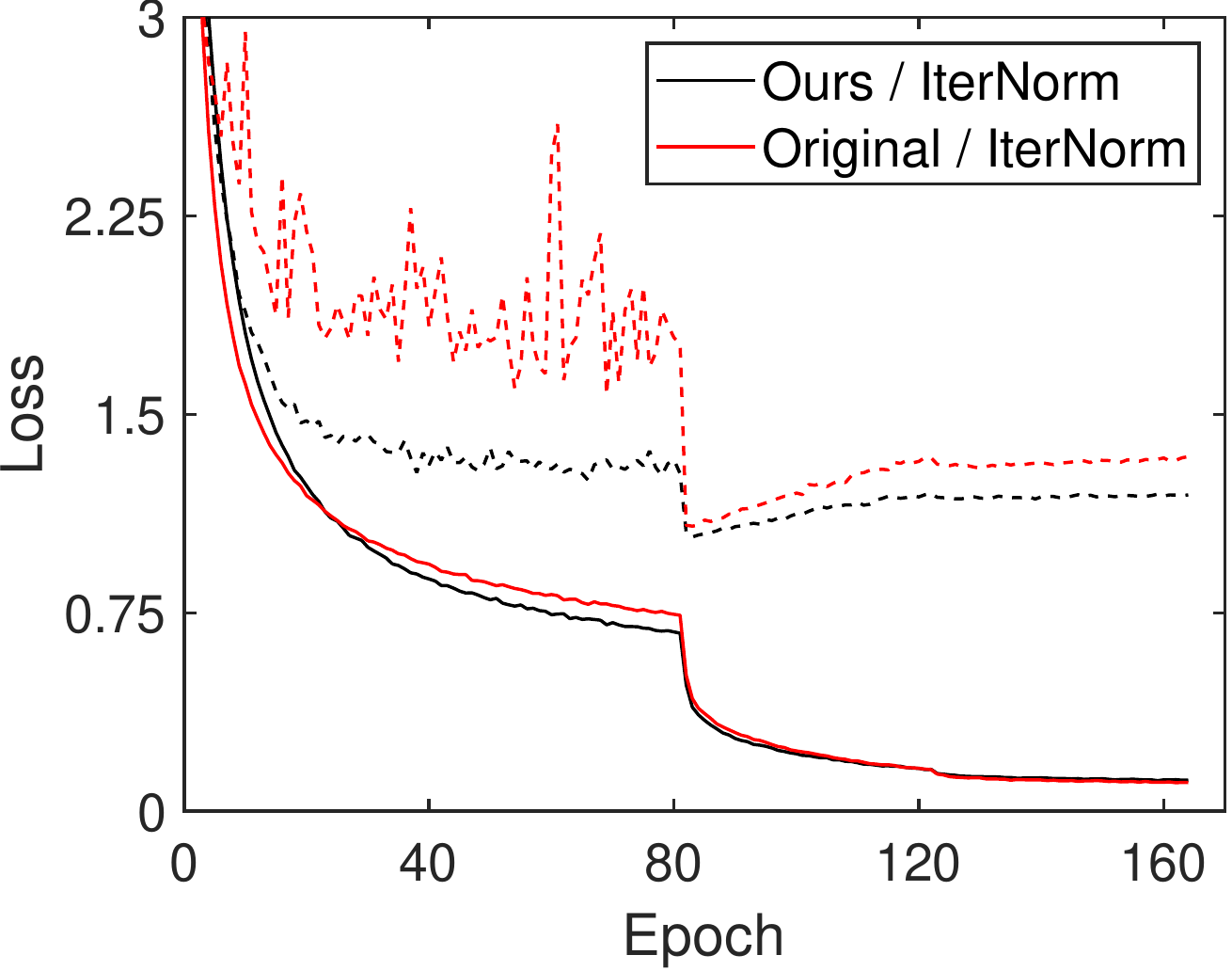}
        \subcaption{Train, Test Loss}
        \label{subfig:loss of mod,basic}
    \end{subfigure}
    \caption{We train ResNet56 on CIFAR-100. The solid line in (a) is the average correlation of convolution input features, and the dashed line in (a) illustrates the average of the rank divided by channel-size of IterNorm input features. (b) is an illustration of train (solid lines) and test (dashed lines) losses of two Convolution Units with respect to epochs.}
    \vspace{-2mm}
    \label{fig:loss,rank,corr of mod,basic}
\end{figure}

Rank of feature is related to the stochasticity of the whitening modules, and stochasticity of the whitening modules is considered as the key property of generalization capability of whitening modules~\cite{huang2020investigation}. If the input feature matrix of the whitening module is not full rank, the output will contain noisy channels caused by stretching the data along the axis with eigenvalues of 0. With the original Convolutional Unit, the input of whitening module, $\boldsymbol{X^l}$ is not full rank when channel-size is increased by point-wise convolution; because, rank of feature is not increased by point-wise convolution, and output of convolution layer is directly connected to whitening module as illustrated in Figure~\ref{subfig:curr}. Moreover, point-wise convolution is commonly used to increase channel-size in practice (e.g. bottleneck block of ResNet~\cite{he2016deep}). By contrast, in our Convolutional Unit, the output of convolution passes the activation function before the whitening module as illustrated in Figure~\ref{subfig:mod1}, and obtains an opportunity to increase its rank. We empirically confirm that the input of the whitening module is full rank in our Convolutional Unit and it is not in the original Convolutional Unit as illustrated in Figure~\ref{subfig:rank,corr}. Our method further enhances stability without controlling iteration number or group size as shown in Figure~\ref{subfig:loss of mod,basic}. It indicates that our Convolutional Unit improves stability of IterNorm without loss of capability of whitening.

\begin{table*}
\begin{center}
\begin{tabular}{l|c|c|c|c|c|c}
\toprule
Dataset & BN & IterNorm & $\gamma,\beta$ & $\gamma$ & $\beta$ & None\\
\toprule
CIFAR-10 & 6.96$\pm$0.12& 7.03$\pm$0.15 & 6.89$\pm$0.26 &7.01$\pm$0.17 & 6.88$\pm$0.06&\textbf{6.66$\pm$0.28} \\
CIFAR-100 &28.37$\pm$0.38 & 28.38$\pm$0.32 & 27.75$\pm$0.12 & 27.83$\pm$0.33 & 27.33$\pm$0.17 & \textbf{27.20$\pm$0.32} \\
\bottomrule
\end{tabular}
\end{center}
\vspace{-3mm}
\caption{Comparisons of test errors ($\%$) on ShiftResNet56 with CIFAR-10/100. All results are computed over 5 random seeds, and shown in the format of ``mean$\pm$std’'. For simplicity, we denote IterNorm using our Convolutional Block by the sort of linear transform and omitting ``Ours / IterNorm''. We denote BN and IterNorm using the original Convolutional Unit with linear transform by ``BN'' and ``IterNorm'' and omitting ``Original / $\gamma, \beta$'', respectively.}
\label{tab:shift+abl}
\end{table*}

\subsection{Shift Operation and Input Misalignment}
\label{subsec:4.shift}
Although Convolutional Unit modification generally improves the performance of IterNorm, BN with original Convolutional Unit still shows better or similar performance in the experiments, and $\beta$ improves performance, despite affecting decorrelation and centering. We assume that the results are caused by the \textit{input misalignment}. Spatial convolution can be generally expressed by eq~\ref{eq:conv eq matrix form}. Spatial convolution spatially shifts input feature before passing to point-wise convolution, and it causes the gap between premise and practice of whitening. We can express \textit{input misalignment} by the following formula:
\vspace{-0.5mm}
\begin{equation}
\label{eq:mis}
    (\boldsymbol{X} \cdot \boldsymbol{X^\top}=\mathbf{I}) \nLeftrightarrow (\boldsymbol{S(X)} \cdot \boldsymbol{S(X)^\top}=\mathbf{I}),
\vspace{-0.5mm}
\end{equation}
where $\boldsymbol{S()}$ is channel-wise shift operation, and $\boldsymbol{X}$ is the input of spatial convolution. $(\boldsymbol{X} \cdot \boldsymbol{X^\top}=\mathbf{I})$ is what whitening modules in Figure~\ref{subfig:mod1} does, and $(\boldsymbol{S(X)} \cdot \boldsymbol{S(X)^\top}=\mathbf{I})$ is what whitening modules in Figure~\ref{subfig:mod2} does. Whitening the input of spatial convolution does not imply whitening the input of point-wise convolution. In order to directly perform whitening at the input of point-wise convolution without modifying the whitening modules, we separate spatial convolution into Grouped Shift~\cite{wu2018shift} and point-wise convolution, and place IterNorm between them. Subsequently, we propose the modified Convolutional Unit that employs Grouped Shift as illustrated in Figure~\ref{subfig:mod2}. For consistency, we employ ShiftResNet as the baseline, which was introduced in~\cite{wu2018shift}, instead of simply replacing the spatial convolution of ResNet with the shift operation and point-wise convolution. We conduct experiments with ShiftResNet as varying the Convolutional Unit and normalization modules on CIFAR-10 and CIFAR-100.

\begin{figure}[t]
    \centering
    \begin{subfigure}[b]{0.23\textwidth}
        \centering
        \includegraphics[width=\textwidth]{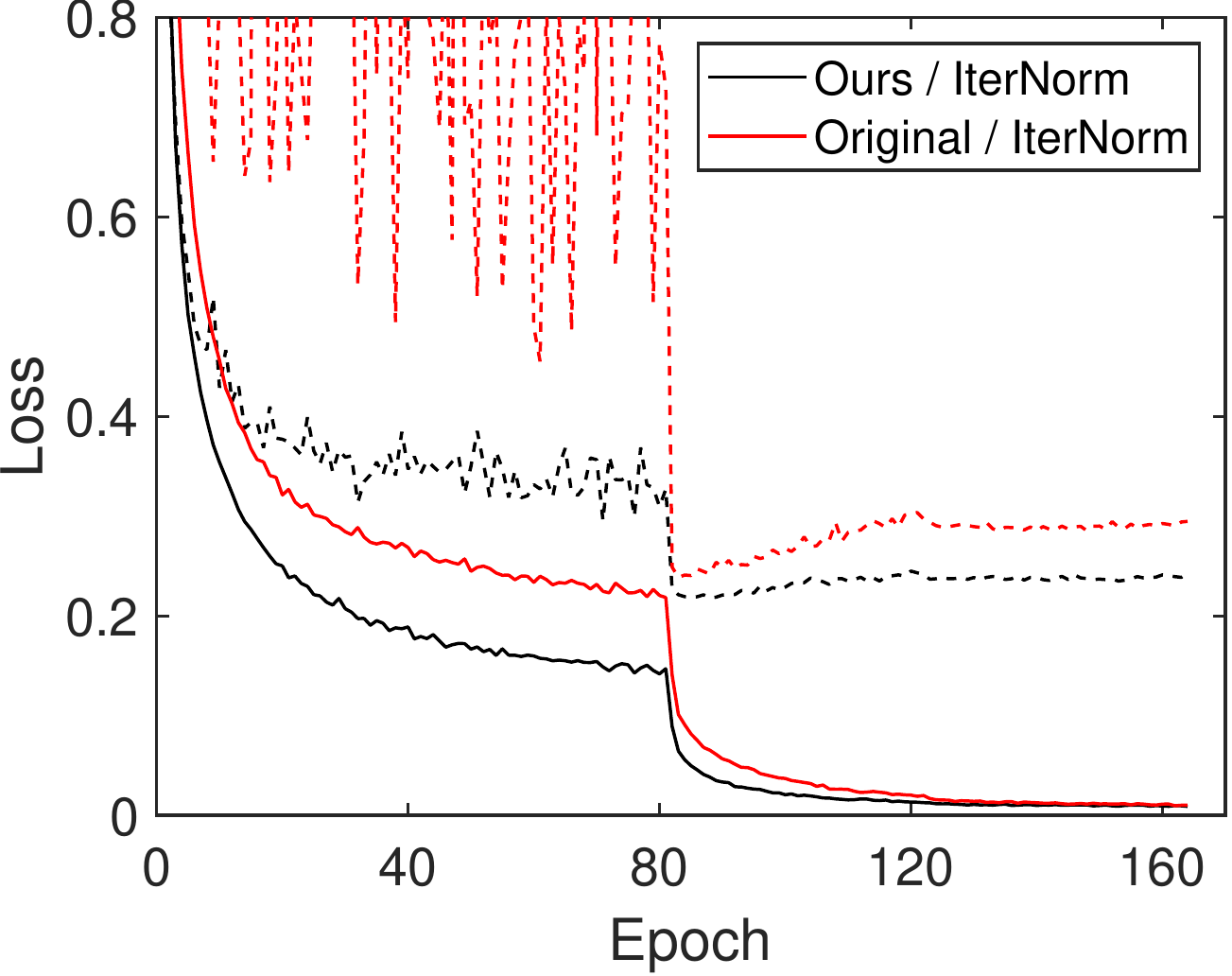}
        \subcaption{CIFAR-10}
    \end{subfigure}
    \hfill
    \begin{subfigure}[b]{0.23\textwidth}
        \centering
        \includegraphics[width=\textwidth]{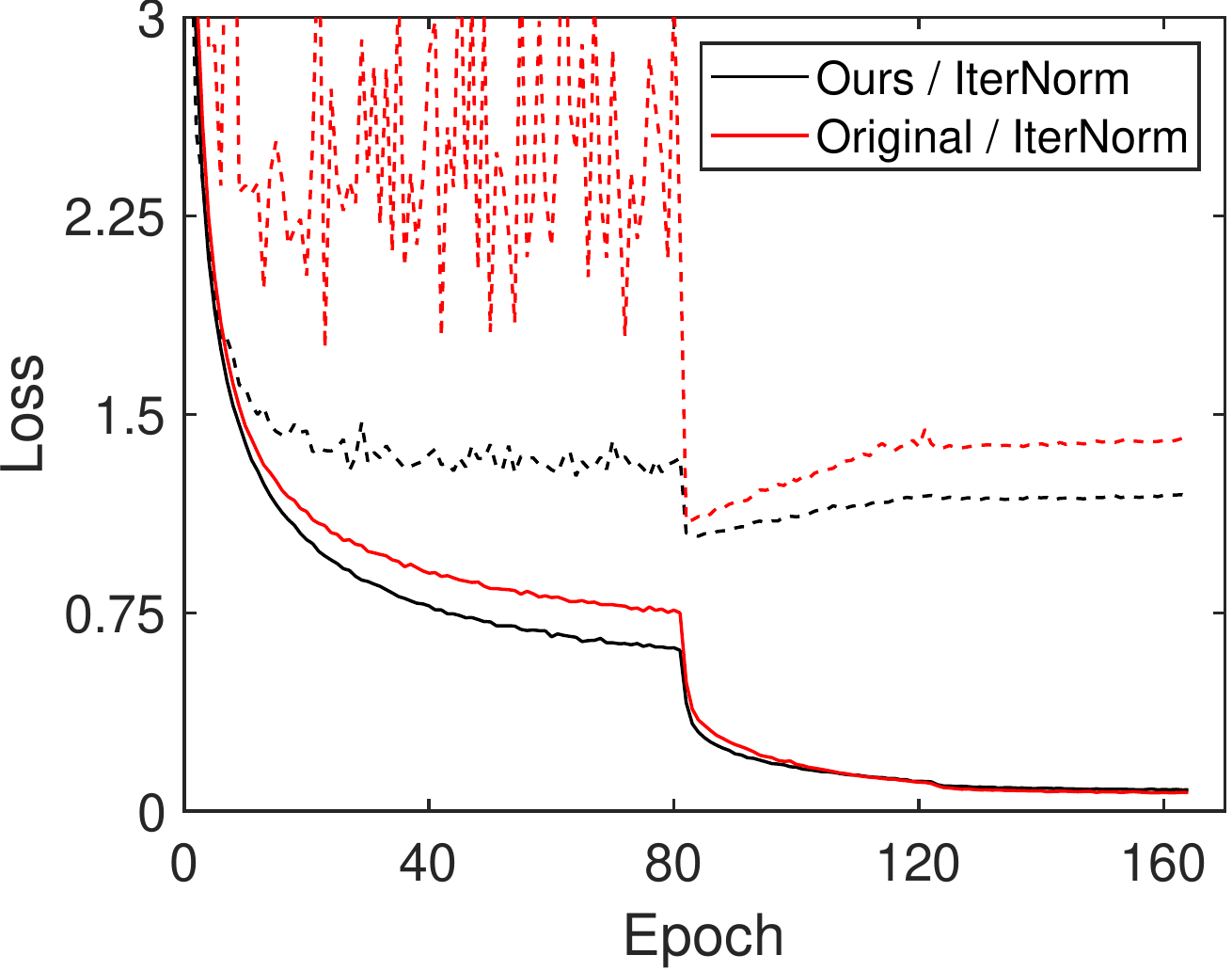}
         \subcaption{CIFAR-100}
    \end{subfigure} 
    \caption{Illustration of train (solid lines), test (dashed lines) loss with respect to epochs. We train ShiftResNet56 by varying the Convolutional Unit on CIFAR-10/100. Both (a) and (b) show that our Convolutional Unit without linear transform improves stability of training.}
    \vspace{-2mm}
    \label{fig:abl_shift+lt}
\end{figure}

From the results shown in Table~\ref{tab:shift+abl}, we verify that the inefficacy of linear transform and original Convolutional Unit, and get general performance improvement as we expected. For simplicity, we do not compare performance by varying linear transform of BN and IterNorm with the original Convolution Unit, because we empirically verified that the original Convolutional Block is well-optimized with linear transform in Section~\ref{sec:lineartransform}. As we assumed, both $\gamma$ and $\beta$ degenerates performance, and IterNorm using our Convolutional Unit shows significantly better performance than both BN and IterNorm using the original Convolutional Unit. By comparing the performance improvement in Table~\ref{tab:abl of base,ours}, we can observe that performance of whitening module is severely affected by \textit{input misalignment} that has not been considered before. From the loss graph in Figure~\ref{fig:abl_shift+lt}, we demonstrate that our Convolutional Unit further improves stability and performance of training. 

\section{Experiment Results}
In this section, we describe details of experiments. We additionally adopt DBN~\cite{huang2018decorrelated} to demonstrate applicability of our Convolutional Unit. DBN is applied with a group size of 64 for our Convolutional Unit, and 16 for original Convolutional Unit, because DBN is highly unstable with a group size larger than 16 with original Convolutional Unit. For simplicity, we call BN, DBN, and IterNorm using the original Convolutional Unit as BN, DBN, and IterNorm, respectively. In the following experiments, we do not use the linear transform for our Convolutional Unit, and we use the linear transform for original Convolutional Unit, unless otherwise stated.

\subsection{Image Classification}

To investigate the effectiveness, we conduct experiments by varying the Convolutional Unit and normalization modules on CIFAR-10, CIFAR-100, CUB-200-2011, Stanford Dogs, and ImageNet~\cite{krizhevsky2009learning,WahCUB_200_2011,KhoslaYaoJayadevaprakashFeiFei_FGVC2011,ILSVRC15}. We demonstrate that our method also improve performance at large-scale dataset and transfer learning. 
\begin{table*}[t]
        \centering
        \tabcolsep=1mm
        \resizebox{0.95\textwidth}{!}{
            \begin{tabular}{@{}l|ccc|ccc@{}}
            \toprule
            & & ShiftResNet 20 & & & ShiftResNet 56 & \\
            \cmidrule{2-7}
            Dataset / Methods & BN & DBN & IterNorm & BN & DBN & IterNorm\\
            \toprule
            CIFAR-10 / Original   &~~ 8.48$\pm$0.27 ~~&~~ 8.95$\pm$0.16 ~~&~~ 8.62$\pm$0.10 ~~&~~ 6.96$\pm$0.12 ~~&~~ 7.42$\pm$0.12 ~~&~~ 7.03$\pm$0.15 ~~ \\
            CIFAR-10 / Ours &~~ - ~~&~~ \textbf{8.43$\pm$0.27} ~~&~~ \underline{8.45$\pm$0.17} ~~&~~ - ~~&~~ \underline{6.84$\pm$0.20} ~~&~~ \textbf{6.66$\pm$0.28} ~~ \\
            \midrule
            CIFAR-10 / Original (lr: 1.0)  &~~ 8.58$\pm$0.37 ~~&~~ 9.40$\pm$0.17 ~~&~~ 9.93$\pm$0.20 ~~&~~ 7.81$\pm$0.09 ~~&~~ 8.17$\pm$0.25 ~~&~~ 8.28$\pm$0.07 ~~ \\
            CIFAR-10 / Ours (lr: 1.0) &~~ - ~~&~~ \underline{7.89$\pm$0.23} ~~&~~ \textbf{7.47$\pm$0.33} ~~&~~ - ~~&~~ \underline{6.33$\pm$0.17} ~~&~~ \textbf{6.04$\pm$0.07} ~~ \\
            \toprule
            CIFAR-100 / Original   &~~ 31.07$\pm$0.40 ~~&~~ 32.52$\pm$0.15 ~~&~~ 31.51$\pm$0.38 ~~&~~ 28.37$\pm$0.38 ~~&~~ 29.87$\pm$0.30 ~~&~~ 28.38$\pm$0.32 ~~ \\
            CIFAR-100 / Ours &~~ - ~~&~~ \underline{30.42$\pm$0.22} ~~&~~ \textbf{30.27$\pm$0.36} ~~&~~ - ~~&~~ \underline{27.85$\pm$0.12} ~~&~~ \textbf{27.20$\pm$0.32} ~~ \\
            \midrule
            CIFAR-100 / Original (lr: 1.0)  &~~ 31.99$\pm$0.49 ~~&~~ 34.12$\pm$0.42 ~~&~~ 34.40$\pm$0.28 ~~&~~ 29.34$\pm$0.34 ~~&~~ 30.85$\pm$0.32 ~~&~~ 30.63$\pm$0.23 ~~ \\
            CIFAR-100 / Ours (lr: 1.0) &~~ - ~~&~~ \underline{29.19$\pm$0.24} ~~&~~ \textbf{29.09$\pm$0.28} ~~&~~ - ~~&~~ \underline{25.71$\pm$0.20} ~~&~~ \textbf{25.51$\pm$0.18} ~~ \\
            \bottomrule
            \end{tabular}
        }
    	\caption{Comparisons of test errors ($\%$) on ShiftResNet 20/56 with CIFAR-10/100. To demonstrate applicability of our method, we additionally employ DBN and get improved performance. Also, to demonstrate enhanced stability of our Convolutional Unit, we train networks with initial learning rate of 1.0 and scheduling by cosine annealing. Best in bold, second-best underlined. All results are computed over 5 random seeds, and shown in the format of ``mean$\pm$std''.}
        \label{tab:cifar,lr}
    \end{table*}
\paragraph{CIFAR-10/100.} For CIFAR datasets, we train the network with 50k training images, and evaluate top-1 errors on 10k test images. Random horizontal flipping and translation by 4 pixels are adopted in our experiments. We use SGD with a batch size of 128 and apply momentum of 0.9 and weight decay of 0.0001. We set the initial learning rate to 0.1, then divide it by 10 at 81 and 122 epochs, and finish the training at 164 epochs. 

From the results shown in Table~\ref{tab:cifar,lr}, performance of DBN and IterNorm with our Convolutional Unit is better than BN, DBN, and IterNorm with original Convolutional Unit. We can observe that performance improvement by our Convolutional Unit increases as depth of network increases. IterNorm using our Convolutional Unit shows $0.37\%$ and $1.18\%$ performance improvement comparing with IterNorm on ShiftResNet56 with CIFAR-10/100, respectively. For DBN, we verify that performance improvement is even larger than IterNorm. As mentioned in~\cite{huang2020investigation, huang2018decorrelated}, DBN suffered from its inherent stochasticity, and our method effectively stabilizes whitening modules by increasing rank of the input matrix. 

\begin{table}[ht]
\begin{center}
\small
    \begin{tabular}{l|c|c}
    \toprule
    Methods & ShiftNet-A & ShiftNet-A-1.5 \\
    \toprule
    BN / Original & 28.81 (9.73) & 23.77 (7.12) \\
    \midrule
    IterNorm / Original & 28.25 (9.50) & 23.87 (7.00)  \\
    \midrule
    IterNorm / Ours & \textbf{27.37 (8.97) } & \textbf{23.15 (6.86)}  \\
    \bottomrule
    \end{tabular}
\end{center}
\vspace{-3mm}
\caption{Comparisons of test errors ($\%$) on ShiftNet-A with ImageNet. To shows performance at deeper and wider networks, we train ImageNet on $\times$1.5 deeper and wider ShiftNet-A. All results are shown in the format of ``top-1 error(top-5 error)''. }
\label{tab:ImgNet}
\end{table} 
\vspace{-3mm}

\paragraph{ImageNet.} We train the network with 1.28M training images and evaluate top-1 and top-5 errors on a validation set with 50k images. We used standard augmentation with 224 pixels cropping. We use SGD with a batch size of 256 and apply a momentum of 0.9 and weight decay of 0.0001. We set the initial learning rate to 0.1, then divide it by 10 at every 30 epochs, and finish the training at 100 epochs.

For consistency, we apply our method on ShiftNet-A that proposed to train ImageNet efficiently in~\cite{wu2018shift}. To verify the effectiveness of our method on larger networks, we compare the performance on 1.5 times wider and deeper ShiftNet-A, which we denote as ``ShiftNet-A-1.5''. For fair comparison, we apply IterNorm using ``Full+DF'' that proposed in~\cite{huang2019iterative} to get the best performance. ``Full+DF'' mean additional IterNorm is applied after last global average pooling. Similar to results on CIFAR, we get the best performance among others. As shown in Table~\ref{tab:ImgNet}, our method obtain $1.44\%$ and $0.62\%$ performance improvement by comparing with BN on ShiftNet-A and ShiftNet-A-1.5, respectively.

\begin{table}[ht]
\begin{center}
\small
\begin{tabular}{l|C|C|C}
\toprule
Dataset & BN & IterNorm & Ours \\
\toprule
CUB& 17.64&16.53 &\textbf{14.10}\\
\midrule
Dogs& 17.75&18.64 &\textbf{16.95}\\
\bottomrule
\end{tabular}
\end{center}
\vspace{-3mm}
\caption{Comparisons of top-1 test errors ($\%$) on ShiftNet-A (pretrained with ImageNet) with CUB-200-2011 and Stanford Dogs. IterNorm with our block shows the best performance on both datasets. }
\label{tab:Transfer}
\end{table}

\paragraph{Transfer Learning.} For CUB-200-2011 and Stanford Dogs, we train with the officially given 5,994 / 12,000 training images, and evaluate top-1 error on 5,794 / 8580 test images, respectively. We use random horizontal flipping with 448 pixels cropping. We use SGD with a batch size of 64 and apply momentum of 0.9 and weight decay of 0.0001. We set the initial learning rate to 0.01, then divide it by 10 at every 30 epochs (15 epochs for Stanford Dogs), and finish the training at 100 epochs (50 epochs for Stanford Dogs). We optimized the configuration by training ShiftNet with BN with a different learning rate in \{0.1, 0.01, 0.001\}. We employ ShiftNet-A pretrained with ImageNet.

As shown in Table~\ref{tab:Transfer}, a significant performance improvement is achieved by our method with computation reduction by removing linear transform. We get $3.54\%$ and $0.8\%$ accuracy improvement comparing with BN on CUB-200-2011 and Stanford Dogs, respectively. To best of our knowledge, it is the first paper that shows applicability of whitening modules in transfer learning. Our results can potentially lead to the future whitening works in transfer learning.

\subsection{Stability}

Next, we domonstrate the superiority of our methods by showing enhanced stability. We train networks with 10 times larger learning rate as varying the Convolutional Unit and normalization layer. We apply cosine annealing~\cite{loshchilov2016sgdr} for a fair comparison without concerning scheduling tuning. As we reported in Table~\ref{tab:cifar,lr}, IterNorm and DBN with our Convolutional Unit shows the best and second best performance among others. Notably, regardless of which whitening module is used, we observe that the performance of whitening module with our Convolutional Unit increases at a larger learning rate unlike with original Convolutional Unit. Comparing with the performance on the basic configuration, IterNorm with our method achieve $0.98\%$ and $0.62\%$ accuracy improvement with CIFAR-10 on ShiftResNet20 and ShiftResNet56, respectively. On CIFAR-100, IterNorm with our method achieve $1.18\%$ and $1.69\%$ accuracy improvement on ShiftResNet20 and ShiftResNet56, respectively. 

\begin{figure}
    \centering
    \begin{subfigure}[b]{0.23\textwidth}
        \centering
        \includegraphics[width=\textwidth]{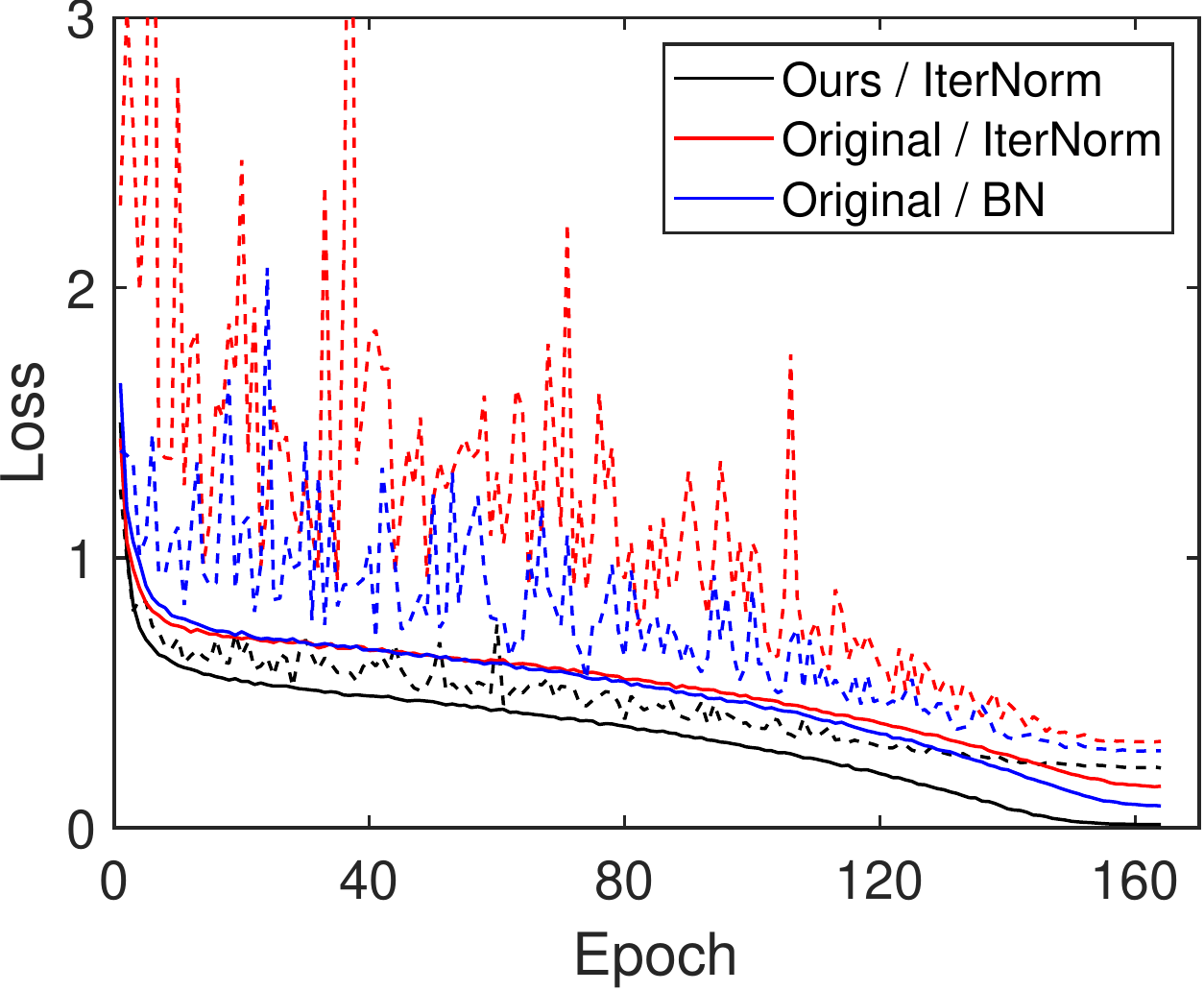}
         \subcaption{CIFAR-10, lr: 1.0}
         \label{subfig:cos_c10}
    \end{subfigure} 
    \hfill
    \begin{subfigure}[b]{0.23\textwidth}
        \centering
        \includegraphics[width=\textwidth]{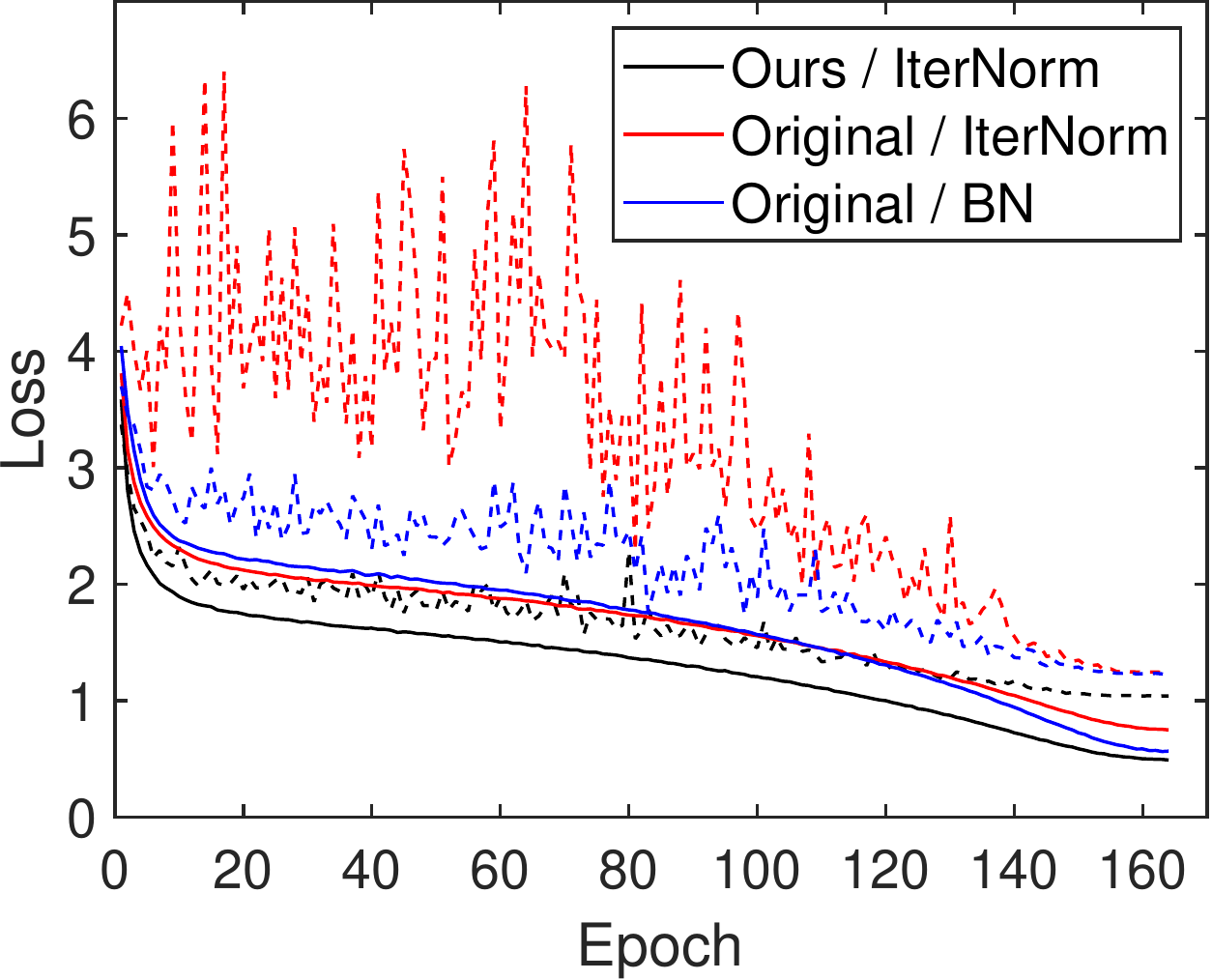}
        \subcaption{CIFAR-100, lr: 1.0}
        \label{subfig:cos_c100}
    \end{subfigure}
    \caption{Illustration of train (solid lines) and test (dashed lines) losses with respect to epochs. We train ShiftResNet20 on CIFAR-10/100 with the initial learning rate of 1.0. We can confirm that IterNorm using the original Convolutional Unit shows extremely unstable test loss. By contrast, IterNorm with our Convolutional Unit shows the best performance and stability among others.}
    \vspace{-2mm}
    \label{fig:cos}
\end{figure}

\begin{figure}[t]
    \centering
    \begin{subfigure}[b]{0.23\textwidth}
        \centering
        \includegraphics[width=\textwidth]{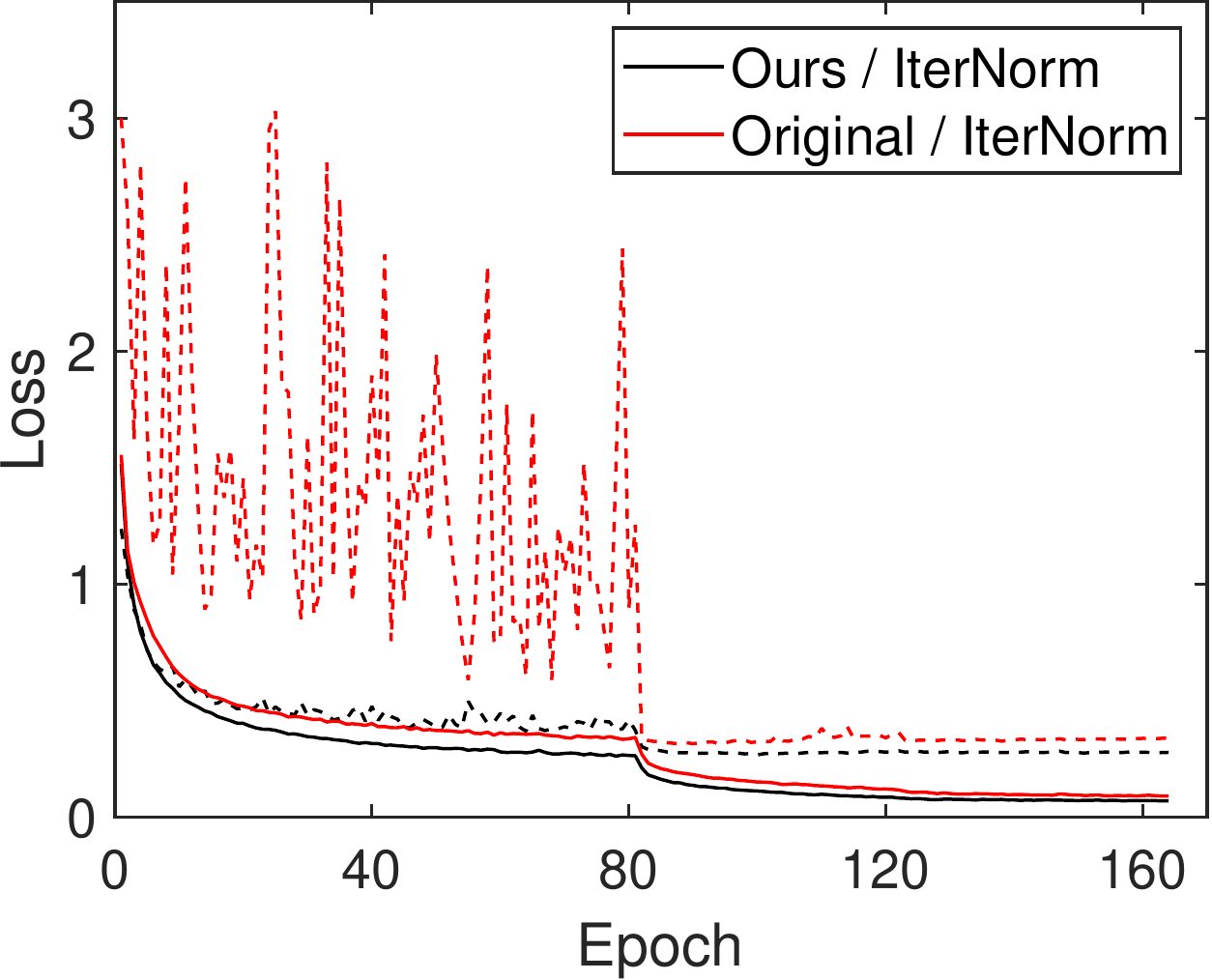}
        \subcaption{CIFAR-10, T: 8}
    \end{subfigure}
    \hfill
    \begin{subfigure}[b]{0.23\textwidth}
        \centering
        \includegraphics[width=\textwidth]{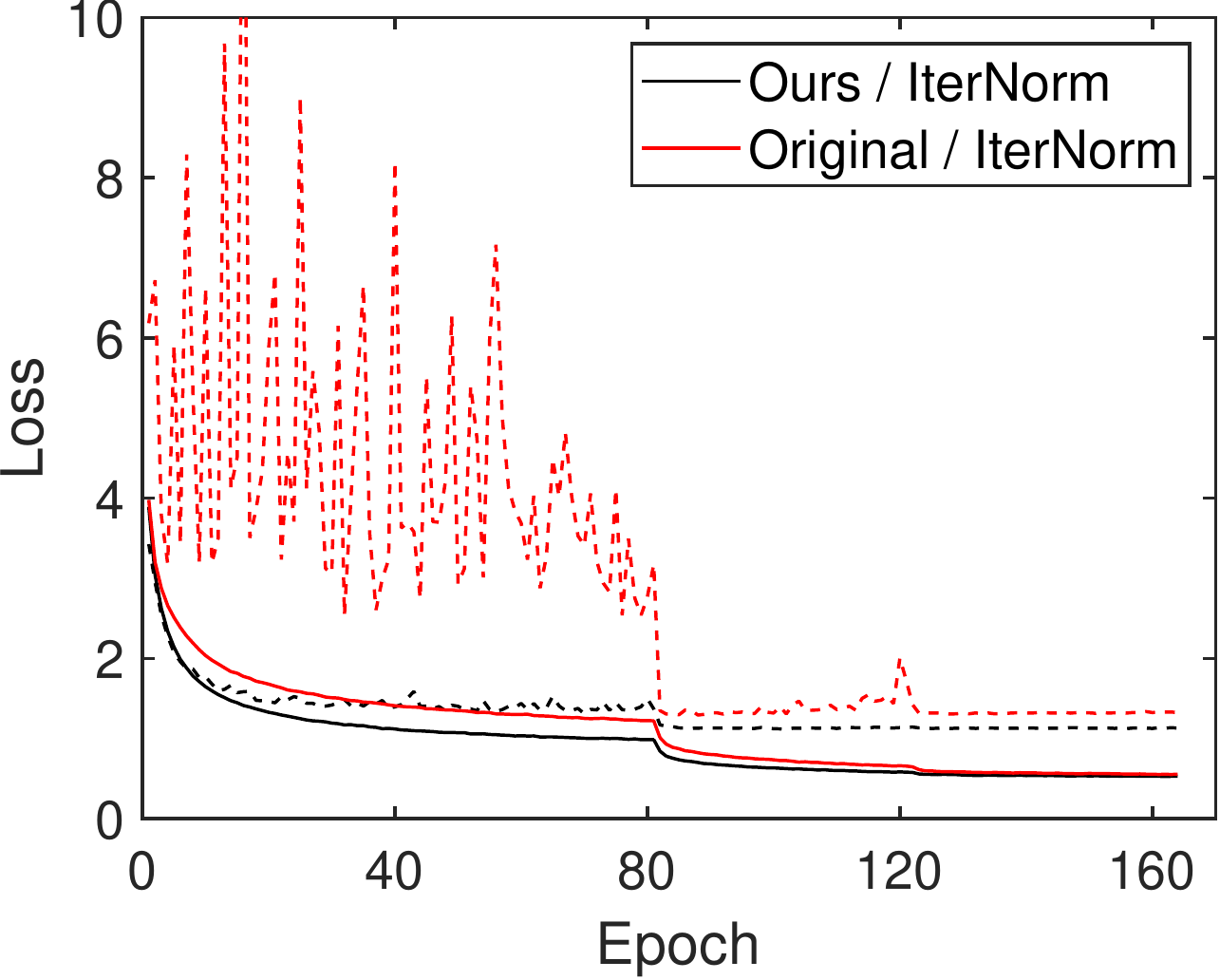}
         \subcaption{CIFAR-100, T: 8}
    \end{subfigure} 
    \caption{Illustration of train (solid lines), test (dashed lines) loss with respect to epochs. We train ShiftResNet20 on CIFAR-10/100 with IterNorm with an iteration number of 8. We can observe that IterNorm using the our Convolutional Unit significantly improves stability when the iteration number is large.}
    \vspace{-2mm}
    \label{fig:T8 loss}
\end{figure}

We also show that our method enhances the stability of whitening modules with large group size and iteration number. Small group size or iteration number reduces stochasticity, but it also reduces whitening capability. IterNorm shows generally good performance at any group size; however, if iteration number is too large, learning become very unstable and shows poor results. This is caused by significant stochasticity due to noisy channels induced by small eigenvalues, and low rank of input feature is investigated in Section~\ref{sec:lineartransform}. Unlike previous studies, our Convolutional Unit fundamentally stabilizes IterNorm by making input features full rank as we confirmed in the Figure~\ref{subfig:rank,corr}. We verify this by comparing the loss graphs of IterNorm with the iteration number of 8. As we have shown in Figure~\ref{fig:T8 loss}, IterNorm using the original Convolutional Unit shows the extremely unstable test loss. By contrast, IterNorm using our Convolutional Unit shows stable test loss, and performance is also similar to that obtained by using the standard iteration number of 5. Also, as mentioned in~\cite{huang2018decorrelated}, DBN with original Convolutional Unit with a group size larger than 16 shows extremely unstable behavior, but DBN with our Convolutional Unit is learnable with full group size.

\section{Conclusion}

In this paper, we investigate the way in which whitening modules, especially IterNorm, can be used. We optimize the efficacy of whitening by bridging the gap between practice and theory of Batch Whitening in terms of block design. The inefficacy of the original Convolutional Unit is empirically investigated, and results are in line with the theory. We demonstrate the improved performance, stability, and transferability of our modified Convolutional Unit, and investigate the correlation and the rank of features to support our results. Our Convolutional Unit significantly stabilizes whitening modules by increasing the rank of features, and improves efficacy by properly choosing the target of whitening and removing the linear transform. Notably, we identify and solve the issue that we denote as \textit{input misalignment}. Without modifying whitening module, we avoid the issue by employing Grouped Shift, and get a significant performance improvement on CIFAR-10/100, CUB-200-2011, Stanford Dogs, and ImageNet. Also, we demonstrate the significantly enhanced stability of our Convolutional Unit at large learning rate, iteration number, and group size.

\clearpage

{\small
\bibliographystyle{ieee_fullname}
\bibliography{egpaper_final}
}

\end{document}